\documentclass[runningheads]{llncs}

\usepackage{eccv}

\usepackage{makecell,multirow}

\usepackage{colortbl}
\definecolor{mygray}{gray}{.9}
\definecolor{mypink}{rgb}{.99,.91,.95}
\definecolor{mycyan}{cmyk}{.3,0,0,0}
\newcommand{\fakeparagraph}[1]{\noindent\textbf{#1}}

\usepackage{eccvabbrv}

\usepackage{graphicx}
\usepackage{booktabs}

\usepackage[accsupp]{axessibility}  

\usepackage{hyperref}

\hypersetup{
    colorlinks = true,
    linkcolor=blue,
    filecolor=green,      
    urlcolor=blue,
    citecolor=green,
}

\usepackage{orcidlink}

\begin{document}

\title{Embracing Events and Frames with Hierarchical Feature Refinement Network for Object Detection} 

\titlerunning{Embracing Events and Frames for Object Detection}

\author{Hu Cao\inst{1}\orcidlink{0000-0001-8225-858X} \and
Zehua Zhang\inst{1}\orcidlink{0009-0007-9220-2441} \and
Yan Xia\inst{1,4}\orcidlink{0000-0001-6684-9814}\and
Xinyi Li\inst{1}\orcidlink{0000-0003-4883-0073}\and
Jiahao Xia\inst{2}\orcidlink{0000-0001-9628-9563}\and
Guang Chen\inst{3}\thanks{Corresponding author}\orcidlink{0000-0002-7416-592X} \and
Alois Knoll\inst{1}\orcidlink{0000-0003-4840-076X}}

\authorrunning{Hu Cao et al.}

\institute{Technical University of Munich, Munich, Germany \\
\email{\{hu.cao,zehua.zhang,yan.xia,super.xinyi,k\}@tum.de}\\  \and
University of Technology Sydney, Sydney, Australia \\
\email{Jiahao.Xia@student.uts.edu.au}\\
\and
Tongji University, Shanghai, China\\
\email{guangchen@tongji.edu.cn} \\
\and
Munich Center for Machine Learning (MCML)}

\maketitle

\begin{abstract}
In frame-based vision, object detection faces substantial performance degradation under challenging conditions due to the limited sensing capability of conventional cameras. Event cameras output sparse and asynchronous events, providing a potential solution to solve these problems. However, effectively fusing two heterogeneous modalities remains an open issue. In this work, we propose a novel hierarchical feature refinement network for event-frame fusion. The core concept is the design of the coarse-to-fine fusion module, denoted as the cross-modality adaptive feature refinement (CAFR) module. In the initial phase, the bidirectional cross-modality interaction (BCI) part facilitates information bridging from two distinct sources. Subsequently, the features are further refined by aligning the channel-level mean and variance in the two-fold adaptive feature refinement (TAFR) part. We conducted extensive experiments on two benchmarks: the low-resolution PKU-DDD17-Car dataset and the high-resolution DSEC dataset. Experimental results show that our method surpasses the state-of-the-art by an impressive margin of $\textbf{8.0}\%$ on the DSEC dataset. 
Besides, our method exhibits significantly better robustness (\textbf{69.5}\% versus \textbf{38.7}\%) when introducing 15 different corruption types to the frame images. The code can be found at the link (https://github.com/HuCaoFighting/FRN).
\end{abstract}


\section{Introduction}

\begin{figure*}[t!]
\centering
\includegraphics[width=\linewidth]{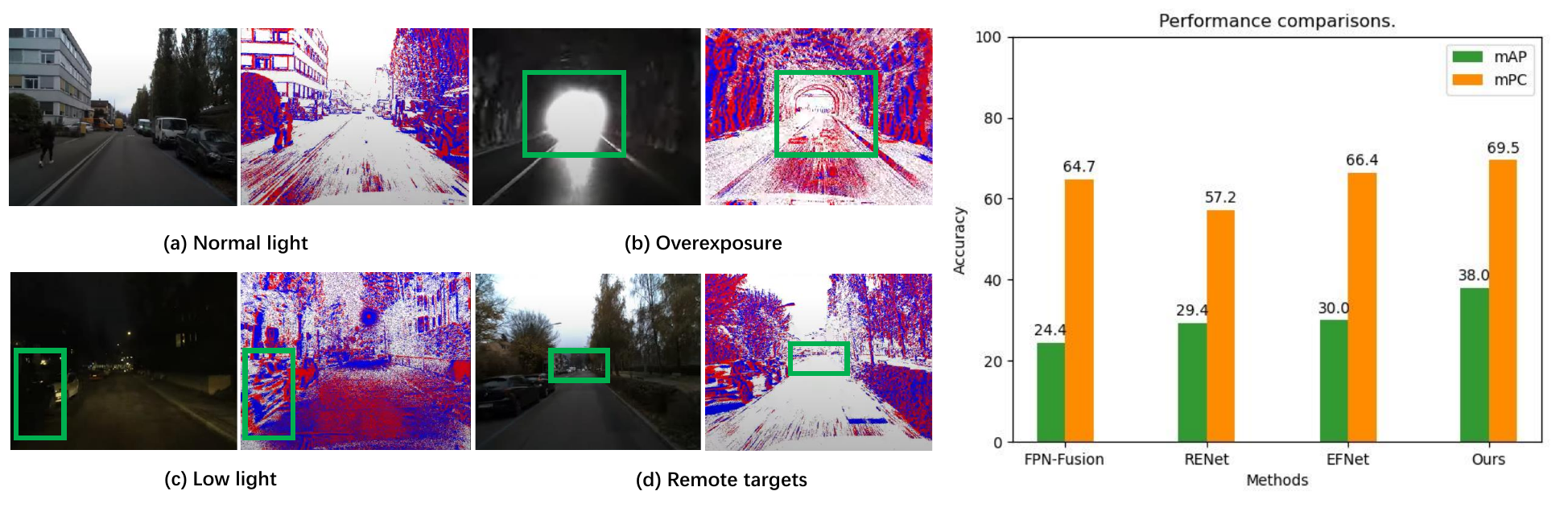}
\caption{This work leverages the complementary information from both events and frames for object detection. (Left) In each image pair, the left image is from frames, while the right is from events. Note that event cameras excel at high-speed and high-dynamic range sensing but struggle to capture static and remote small targets compared to RGB cameras. 
(Right) We choose three methods, FPN-Fusion~\cite{FPN_Fusion}, RENet~\cite{zhou2023rgb}, and EFNet~\cite{sun2022event} for performance evaluation.} 
\label{examples}
\end{figure*}

Object detection is a fundamental task in computer vision~\cite{oksuz2020imbalance,liu2020deep, auin}. The performance of conventional frame-based cameras in object detection often faces a significant decline in challenging conditions, such as high-speed motion, poor lighting conditions (e.g. low-light and overexposure), and image corruptions~\cite{sun2019fab,hu2020learning,michaelis2019benchmarking}.
Recently, an emerging bio-inspired vision sensor, the event camera (e.g., dynamic and active pixel
vision sensors, DAVIS), has been developed for vision perception~\cite{brandli2014240, moeys2017sensitive}. The working principle of the event camera fundamentally differs from that of the conventional camera. It transmits only the local pixel-level changes caused by variations in lighting intensity, like a bio-inspired retina~\cite{chen2020event, gallego2020event}. The event camera has several appealing characteristics, including low latency, high dynamic range (120 dB), and high temporal resolution. It offers a new perspective to address some limitations of conventional cameras in challenging scenarios. However, similar to the problems faced by conventional cameras in extreme-light scenarios, event cameras exhibit poor performance in static or remote scenes with small targets, as illustrated in Fig.~\ref{examples}. The event camera captures dynamic context and structural information, whereas the frame-based camera provides rich color and texture information. Both event cameras and frame-based cameras are complementary, motivating the development of new algorithms for various computer vision tasks.

\begin{figure}[t!]
\centering
\includegraphics[width=\linewidth]{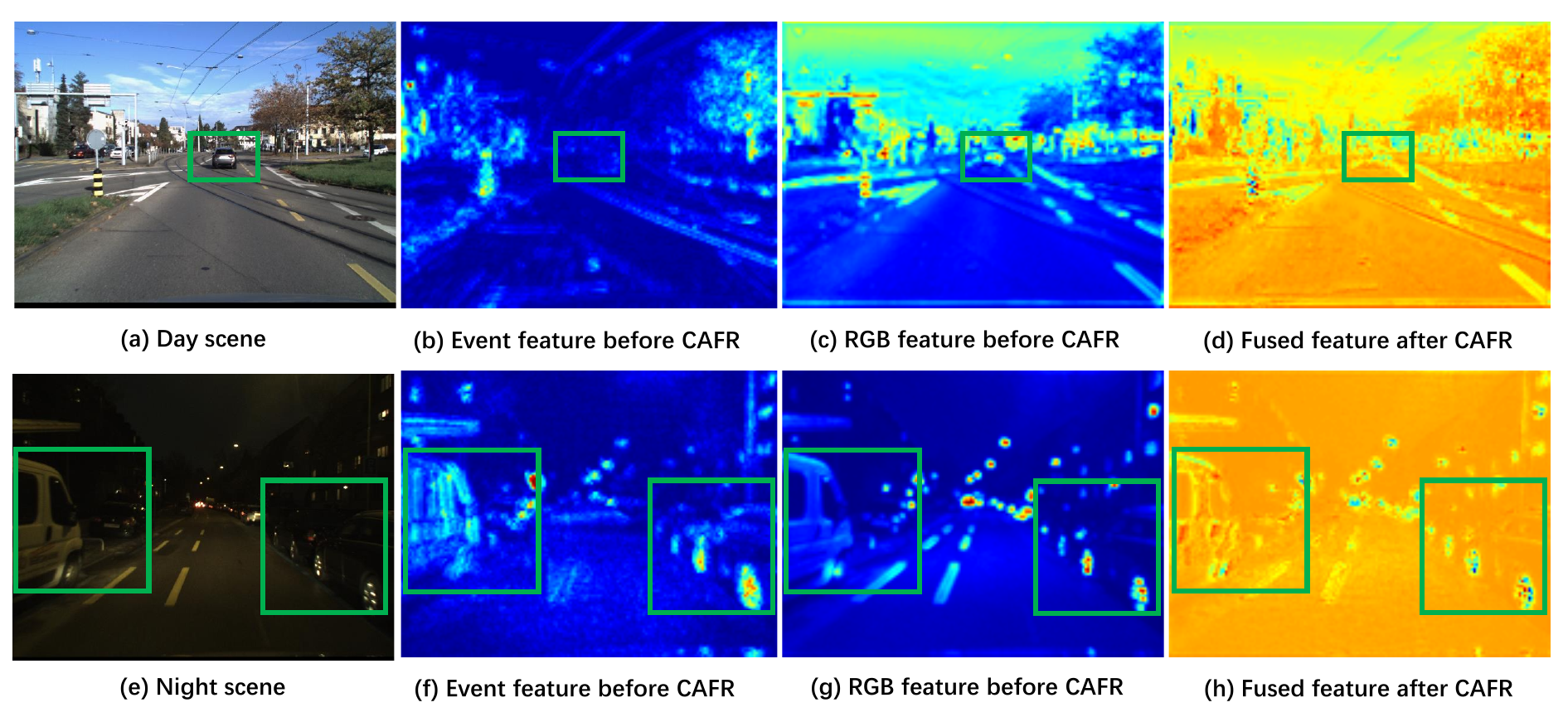}
    \caption{Feature maps of RGB and event modalities before and after CAFR. The first row corresponds to the day scene, and the last row represents the night scene.} 
\label{FeatureMap}
\end{figure}

Current event-frame fusion methods often utilize concatenation~\cite{FPN_Fusion,neurograsp}, attention mechanisms~\cite{FAGC,zhou2023rgb,sun2022event}, and post-processing~\cite{jiang2019mixed,multi_cue,JDF} strategies to fuse events and frames. Simply concatenating event-based and frame-based features improves performance slightly, but the inherent complementary nature between different modalities is not fully exploited. The authors of~\cite{FAGC} employ pixel-level spatial attention to leverage event-based features in enhancing frame-based features, leading to improved performance. However, enhancing single-modal features is considered suboptimal, as event-based features and frame-based features possess unique characteristics. The methods presented in~\cite{zhou2023rgb,sun2022event} incorporate feature interaction between event-based and frame-based features; however, they do not comprehensively consider the feature imbalance problems existing in event-frame object detection.
As illustrated in Fig.~\ref{FeatureMap}, we present the feature maps before and after the application of our fusion module. In the daytime scene, event-based features struggle to effectively capture the remote car compared to RGB-based features. Conversely, in the night scene, RGB-based features face challenges in capturing cars effectively compared to event-based features due to lighting conditions. The feature-modality imbalance problem stems from the misalignment and inadequate integration of different modalities. Addressing this issue requires fully incorporating cross-modal complementarity to generate robust features.

To address the aforementioned problems, we propose a novel hierarchical feature refinement network with CAFR modules for event-frame fusion. In contrast to the current event-frame fusion methods, our method adopts a dual-branched coarse-to-fine structure. The dual-branch architecture guarantees comprehensive utilization of both event-based and frame-based features. To extract refined features, following the initial feature interaction, we further employ feature statistics to align features at the channel-level mean and variance, thereby enhancing the model's representation capabilities. Specifically, CAFR consists of two parts: the bidirectional cross-modality interaction (BCI) part and the two-fold adaptive feature refinement (TAFR) part, where attention mechanisms and feature statistics are utilized to balance feature representations. Extensive experiments demonstrate that our method outperforms the state-of-the-art (SOTA) on several datasets (PKU-DDD17-Car, DSEC, and corruption data).

The main contributions of this work can be summarized as follows:   
\begin{itemize}
\item We propose a novel hierarchical feature refinement network with CAFR modules to fuse events and frames, which enriches extracted features with more valuable information, thereby enhancing the overall detection performance.
\item BCI and TAFR parts are devised to constitute the CAFR by incorporating sufficiently complementary information to form discriminative representations for robust object detection.
\item Our method outperforms SOTA methods on the low-resolution PKU-DDD17-Car dataset, the high-resolution DSEC dataset, and the corruption data. As shown in Fig.~\ref{examples}, our model outperforms the second-best model, EFNet~\cite{sun2022event}, by an impressive margin of $\textbf{8.0}\%$ mAP and $\textbf{3.1}\%$ mPC, respectively.
\end{itemize}
\section{Related Work}

\fakeparagraph{Event-based object detection.}
Several event-based detection methods, such as RED~\cite{perot2020learning}, ASTMNet~\cite{ASTMNet}, and RVT~\cite{gehrig2023recurrent}, leverage the remarkable capabilities of event cameras. RED~\cite{perot2020learning} introduces a ConvLSTM recurrent network architecture for extracting spatio-temporal features from event streams. Similarly, ASTMNet~\cite{ASTMNet} introduces a spatio-temporal attentional convolution module for learning event feature embeddings, alongside a lightweight spatio-temporal memory module for extracting cues from continuous event streams. On the other hand, RVT~\cite{gehrig2023recurrent} proposes a pioneering backbone algorithm for object detection, substantially reducing inference time while maintaining performance levels comparable to previous methods. Recently, state-space models (SSMs) were introduced to improve the generalizability of event-based object detection~\cite{Zubic_2024_CVPR}. Nevertheless, the absence of color and fine-grained texture details in event streams results in insufficient semantic information, which is crucial for effective detection tasks.

\fakeparagraph{Event-frame fusion for object detection.}
In the field of object detection, the authors of~\cite{JDF} and ~\cite{multi_cue} adopt the Dempster-Shafer theory to fuse events and frames for vehicle detection and pedestrian detection, respectively. Additionally, Chen et al.~\cite{frequency} employ non-maximum suppression (NMS) for fusing the detection outcomes of two modalities, while Jiang et al.~\cite{jiang2019mixed} suggest fusing confidence maps derived from the two modalities. All the above methods can be categorized as late fusion methods. While late fusion can be effective, these methods often lack feature interactions and struggle to fully exploit the complementary nature of the fused modalities. Recently, various middle fusion methods~\cite{FPN_Fusion,FAGC,zhou2023rgb} have been proposed to guide the fusion of two modalities at the feature level. In~\cite{FPN_Fusion,neurograsp}, simple concatenation is employed to fuse event-based and frame-based features for performance improvement. The authors of~\cite{FAGC} utilize pixel-level spatial attention to leverage event-based features for enhancing frame-based features, resulting in improved performance. However, due to the uniqueness of event-based features and frame-based features, enhancing single-mode features is considered suboptimal. The approach introduced in~\cite{zhou2023rgb} incorporates feature interaction between event-based and frame-based features, but they do not fully address the feature imbalance issues present in event-frame object detection.

\fakeparagraph{Various fusion modules for events and frames.}
Multimodal fusion methods for events and frames have been explored across various vision tasks, including deblurring~\cite{sun2022event}, steering angle prediction~\cite{munir2023multimodal}, depth estimation~\cite{gehrig2021combining}, and semantic segmentation~\cite{CMX}. In~\cite{sun2022event}, the cross-self-attention is proposed as a fusion module to combine events and frames to improve deblurring performance. In contrast, the authors of~\cite{munir2023multimodal} processed events and frames separately with the self-attention module and then summed the outputs to get the fusion features. For depth estimation, RAM Net is developed in~\cite{munir2023multimodal} to leverage events and frames' information. Moreover, a unified fusion framework, CMX, is introduced in~\cite{CMX} for RGB-X semantic segmentation.

\section{Method}
In this section, we first introduce preliminaries, including the working principles of the event camera and event representation. Subsequently, we delve into a detailed illustration of our proposed hierarchical feature refinement network.

\subsection{Preliminaries}
\fakeparagraph{Event camera.} The conventional frame-based camera operates by capturing and delivering a sequence of frames at a fixed frequency. In stark contrast, the working principle of the event camera deviates fundamentally from that of its frame-based counterpart. The event camera, a bio-inspired vision sensor, generates asynchronous and sparse event streams exclusively when the change in logarithmic intensity $L(x,y,t)$ surpasses a predetermined threshold $C$. The computational process can be succinctly formulated as follows:

\begin{equation}
\begin{aligned}
L(x,y,t) - L(x,y,t-\Delta t) \geq pC, \quad p \in\left\{-1,+1\right\}.\\
\end{aligned}
\end{equation}
where $\Delta t$ represents the time variation. The event polarity, denoted by $p \in\left\{-1,+1\right\}$, signifies the sign of the brightness change, indicating either positive (``ON'') or negative (``OFF'') events, respectively. The generated event streams $E$ can be expressed as follows:

\begin{equation}
	E = \left\{e_i\right\}_{i=1}^N, e_i = (x_i, y_i, t_i, p_i).
	\label{eq:event}
\end{equation}
where $N$ denotes the number of events $e_i$ within the event stream $E$. The tuple $(x,y)$ denotes the triggered pixel coordinates, while $t$ and $p$ represent the corresponding timestamp and polarity, respectively.

\fakeparagraph{Event representation.}
\label{voxel_grid}
The event representation employed in this study is the voxel grid~\cite{zhu2019unsupervised} generated through the discretization of the time domain. Considering a set of \(N\) input events \(\{(x_i, y_i, t_i, p_i)\}_{i \in [1, N]}\) and a set of \(B\) bins for discretizing the time dimension, we normalize the timestamps to the range \([0, B - 1]\). Subsequently, the event volume is generated as follows:

\begin{equation}
\begin{aligned}
t^*_i &= \frac{(B - 1)(t_i - t_1)}{t_N - t_1}, \\
V(x, y, t) &= \sum_i p_i k_b(x - x_i) k_b(y - y_i) k_b(t - t^*_i), \\
k_b(a) &= \max(0, 1 - |a|).
\end{aligned}
\label{equation_VoxelGrid}
\end{equation}
where \(k_b(a)\) corresponds to the bilinear sampling kernel, as defined in~\cite{redmon2016you}. More specifically, we regard the time domain of the voxel grid as channels in a conventional 2D image and conduct a 2D convolution across the spatial dimensions $x$ and $y$. This approach enables the model to effectively capture feature representations from the spatiotemporal distribution of events.

\subsection{Hierarchical Feature Refinement Network}

\begin{figure}[t!]
\centering
\includegraphics[width=0.95\linewidth]{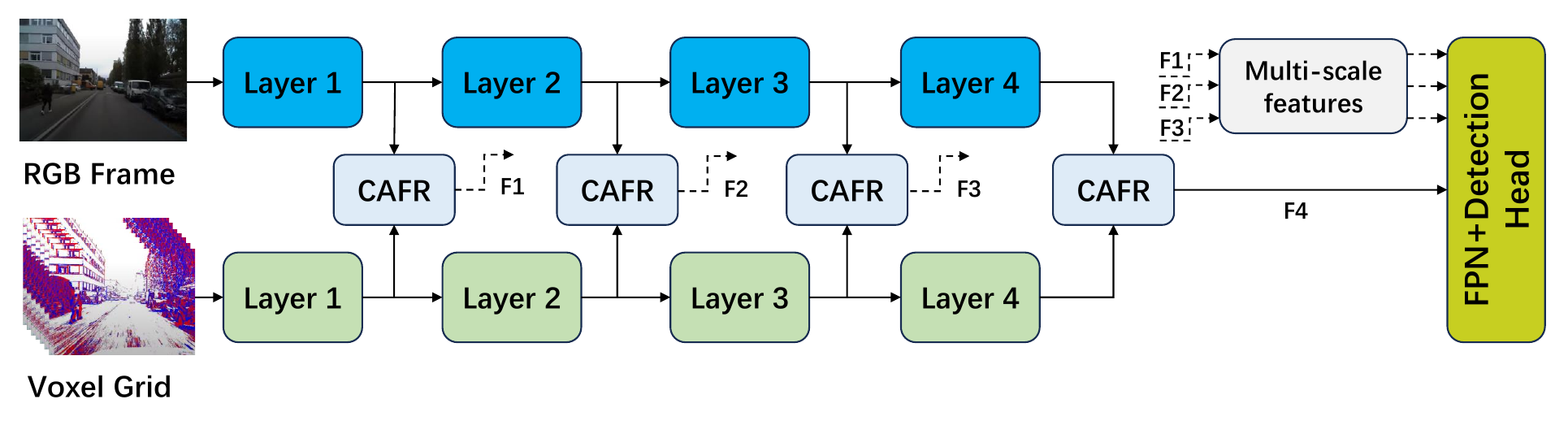}
\caption{The overall architecture of our hierarchical feature refinement network. It comprises a dual-stream backbone network, CAFR, FPN, and a detection head. The backbone incorporates two branches: the event-based ResNet-50 (bottom) and the frame-based ResNet-50~\cite{resnet} (top). The CAFR operates to enhance features on a hierarchical scale. The refined multi-scale features are then forwarded to the FPN and detection head for accurate detection predictions. The structure of the FPN and the detection head is adapted from~\cite{RetinaNet}.} 
\label{FRN}
\end{figure}

\fakeparagraph{Overview.}
The method's architecture, illustrated in Fig.~\ref{FRN}, comprises a dual-stream backbone, CAFR, a feature pyramid network (FPN), and a detection head. Events are preprocessed using the voxel grid encoding method for CNN to extract deep features. Utilizing event-frame data as inputs, the dual-stream backbone network extracts multi-scale features. For effective information exchange between different modal features, CAFR receive event-based and frame-based features to balance the information flow. Subsequently, the detection head operates on the harmonized multi-modal features for precise detection predictions. Detailed explanations of each module will be provided in the subsequent parts.

\fakeparagraph{Dual-stream backbone.}
The backbone incorporates two branches: event-based and frame-based ResNet-50s~\cite{resnet}. Each ResNet-50 is composed of four blocks denoted as $\left\{L_1, L_2, L_3, L_4\right\}$. The feature maps' resolution progressively decreases from $L_1$ to $L_4$, with the resolution of the features being consistently maintained at each block. Residual learning is employed to extract semantically stronger and more valuable features. In this work, a CAFR module is strategically inserted between the two blocks to enhance the learning process.

\begin{figure}[t!]
\centering
\includegraphics[width=0.9\linewidth]{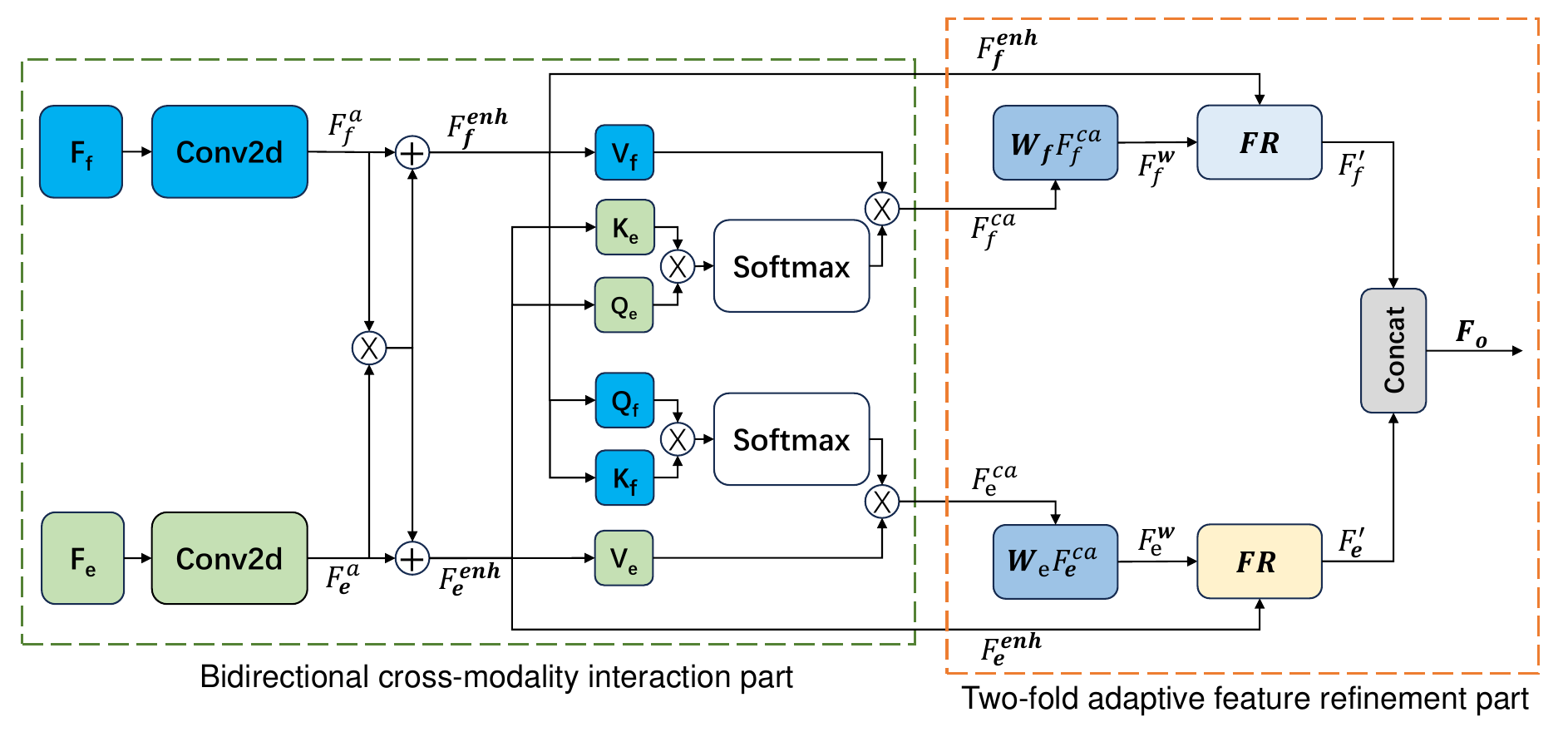}
\caption{Cross-modality adaptive feature refinement module (CAFR). It contains two integral parts: bidirectional cross-modality interaction (BCI) and two-fold adaptive feature refinement (TAFR). Here, ``FR'' denotes feature refinement.} 
\label{AFRM}
\end{figure}

\fakeparagraph{Cross-modality adaptive feature refinement (CAFR) module.}
Frame-based cameras perform poorly in several challenging scenarios, such as overexposure, high-speed motion, etc. Furthermore, frame-based features may generate unclear results because objects share many visual similarities. In contrast, event cameras with high temporal resolution and dynamic range excel at capturing motion and edge information. Recognizing the complementarity of event-based and frame-based information, we introduce the CAFR to enhance feature representations at the feature level by leveraging event-based dynamic context.
As depicted in Fig.~\ref{AFRM}, CAFR processes both frame-based features $F_{f}$ and event-based features $F_{e}$ to obtain balanced semantic features. 

\emph{Bidirectional cross-modality interaction (BCI).} Initially, a transformation module, utilizing a $1\times 1$ convolution layer, is employed for activation. The calculation can be expressed as follows:

\begin{equation}
\begin{aligned}
F_{f}^a = \text{Conv}_{1 \times 1}(F_{f}), F_{e}^a = \text{Conv}_{1 \times 1}(F_{e}). 
\end{aligned}
\end{equation}

Subsequently, the activated frame and event features undergo a coarse-to-fine fusion. Specifically, a global attention map is computed through pixel-wise multiplication, representing mixed attention across spatial and channel dimensions to enhance features. Formally, the computation for enhanced features $F_{f}^{enh}$ and $F_{e}^{enh}$ is as follows:

\begin{equation}
\begin{aligned}
F_{f}^{enh} = F_{f}^a \otimes F_{e}^a + F_{f}^a, F_{e}^{enh} = F_{f}^a \otimes F_{e}^a + F_{e}^a.
\end{aligned}
\end{equation}

To bridge information from two distinct sources, the features undergo processing via a bidirectional cross-self-attention mechanism at a coarse level. The input features $F_{f}^{enh}$ and $F_{e}^{enh}$ are first projected into query ($Q_f$ and $Q_e$), key ($K_f$ and $K_e$), and value ($V_f$ and $V_e$) tensors. The computation can be expressed as follows:

\begin{equation}
\begin{aligned}
Q_f&=F_{f}^{enh}W_f^Q, K_f=F_{f}^{enh}W_f^K,   \\
V_f&=F_{f}^{enh}W_f^V, Q_e=F_{e}^{enh}W_e^Q, \\
K_e&=F_{e}^{enh}W_e^K, V_e=F_{e}^{enh}W_e^V. \\
\end{aligned}
\end{equation}
where $W$ represents a learnable linear projection.

In contrast to self-attention in~\cite{10584449,xia2021soe,xia2023casspr}, which focuses on relationships within an individual feature modality, cross-self-attention extends its scope by incorporating guidance from the counterpart feature modality. This mechanism captures semantics as:

\begin{equation}
\begin{aligned}
F_{f}^{ca} = CrossAtt_f(Q_{e},K_{e},V_{f})&, F_{e}^{ca} = CrossAtt_e(Q_{f},K_{f},V_{e}). \\
\end{aligned}
\end{equation}
Where $CrossAtt_f$ is defined as follows:

\begin{equation}
\begin{aligned}
CrossAtt_f(Q_{e},K_{e},V_{f}) &= SoftMax(\frac{Q_{e}K_{e}^T}{\sqrt{D}})V_{f}. \\
\end{aligned}
\end{equation}
where $D$ denotes the channel dimension of the feature map. $CrossAtt_e$ is computed in a similar manner.

\emph{Two-fold adaptive feature refinement (TAFR).} 
Furthermore, we employ feature statistics to align the feature space, leveraging insights from the style transfer domain~\cite{huang2017arbitrary}. Specifically, the features undergo further refinement by aligning the channel-level mean and variance. The resulting final output, denoted as $F_{o}$, is attained by concatenating the two aligned features. This entire process is represented as follows:

\begin{equation}
\begin{aligned}
F_{f}^{w} &= W_{f}F_{f}^{ca}, \quad F_{e}^{w} = W_{e}F_{e}^{ca}, \\
F^{'}_{f} &= \frac{\sigma(F_{f}^{enh})(F_{f}^{w} - \mu(F_{f}^{w}))}{\sigma(F_{f}^{w})} + \mu(F_{f}^{enh}),\\
F^{'}_{e} &= \frac{\sigma(F_{e}^{enh})(F_{e}^{w} - \mu(F_{e}^{w}))}{\sigma(F_{e}^{w})} + \mu(F_{e}^{enh}),\\
F_{o} &= Concat(F^{'}_{f},F^{'}_{e}).\\
\end{aligned}
\end{equation}
where $W_{f}$ and $W_{e}$ represent learnable linear projections.
Notably, the computation for $F^{'}_{f}$, $F^{'}_{e}$, and $F_{o}$ operates without learnable affine parameters. Instead, it adaptively computes the affine parameters directly from the inputs. The computation of \( \mu \) and \( \sigma \) is expressed as:

\begin{equation}
\begin{aligned}
\mu(x) &= \frac{1}{H \times W} \sum_{h=1}^{H} \sum_{w=1}^{W} x, \\
\sigma(x) &= \sqrt{\frac{1}{H \times W} \sum_{h=1}^{H} \sum_{w=1}^{W} (x - \mu(x))^2 + \epsilon}.
\end{aligned}
\end{equation}
Where $x$ denotes the input feature maps. \( H \) and \( W \) represent the height and width of the input feature map, respectively. \( \epsilon \) is a small value introduced to prevent division by zero, typically set to \( 1 \times 10^{-5} \).

\fakeparagraph{FPN and detection head.}
Similar to the previous works~\cite{RetinaNet,fpn,FAGC}, we employ FPN to enhance features, improving detection robustness for objects of varying sizes. The outputs $\left\{P_{n}\right\}_{n=1}^4$ are generated through top-down pathways and lateral connections. The last level of the feature map, $P_{5}$, is obtained by applying a $3\times3$ convolutional layer with stride 2 on $P_{4}$. The resulting multi-level feature maps $\left\{P_{n}\right\}_{n=1}^5$ are then fed into the detection head for prediction. 

After FPN processing, the detection head comprises two subnets responsible for classification and bounding box regression. Each sub-network comprises four $3\times 3$ convolutional layers with 256 filters. In the classification sub-network, followed by a $3\times 3$ convolutional layer with $KA$ filters, followed by sigmoid activations, it outputs $KA$ binary predictions. In the bounding box regression sub-network, followed by a $3\times 3$ convolutional layer with $4A$ filters, it outputs $4A$ offset predictions. $A$ is set to $9$. The specific offset parameters of the bounding box can be expressed as follows:

\begin{equation}
	\label{equation_2}
	\begin{aligned}
		t_x^{'} &= \frac{(x^{'} - x_a)}{w_a}, t_y^{'}  = \frac{(y^{'}  - y_a)}{h_a},\\
		t_w^{'}  &= log(\frac{w^{'}}{w_a}),\quad t_h^{'}  = log(\frac{h^{'}}{h_a}).\\
	\end{aligned}
\end{equation}
where $x, y, w, h$ denote the center coordinates, width, and height of the bounding box, respectively. The variables $t^{'}, x^{'}, x_a$ represent the prediction regression offsets, predicted bounding box, and anchor box, respectively.

\section{Experiments}
In this section, we outline the datasets and evaluation metrics used for event-frame multimodal object detection. Extensive experiments are then conducted to evaluate the effectiveness of our proposed method.

\subsection{Datasets}

\fakeparagraph{DDD17.}
The DDD17 dataset~\cite{DDD17} is collected to combine events and frames for end-to-end driving applications. It uses a $346\times 260$ pixel DAVIS to record over 12 hours of highway and city driving in various conditions (daytime, evening, night, dry, etc.). However, the original DDD17 dataset does not have object detection labels. The authors of~\cite{JDF} manually labeled the vehicles in the DDD17 dataset to make them available for object detection tasks, named PKU-DDD17-Car. The PKU-DDD17-Car dataset contains frame data and corresponding event streams, of which 2241 frames are the training set and 913 frames are the test set.

\fakeparagraph{DSEC.}
DSEC~\cite{gehrig2021dsec} is a high-resolution and large-scale event-frame dataset for real-world driving scenarios. 
Concurrently, event data is captured using an event camera with a resolution of 640 × 480. Unlike the DDD17 dataset, where both event data and frame data originate from the same DAVIS camera, the DSEC dataset features non-synchronized event and RGB data. To address this, a homographic transformation based on camera matrices is employed to align the viewpoint and resolution of the event and RGB cameras. The initial release of the DSEC dataset lacks annotations for object detection. In this study, we utilize the labels introduced in~\cite{FPN_Fusion} for evaluation. Specifically, the labeled dataset encompasses three object categories: cars, pedestrians, and large vehicles.

\fakeparagraph{Corruption data.}
Previous studies~\cite{michaelis2019benchmarking,dodge2016understanding,FPN_Fusion} indicate significant performance degradation (as low as 30–60$\%$ of the original performance) in standard object detection models when applied to corrupted images. To assess model's robustness, we introduced 15 distinct corruption types to frame images, including gaussian noise, shot noise, impulse noise, defocus blur, frosted glass blur, motion blur, zoom blur, snow, frost, fog, brightness, contrast, elastic, pixelate, and JPEG compression. These corruptions can be broadly categorized into four groups: noise, blur, weather, and digital. 
Each corruption type is subjected to five severity levels. Please find more details in the \textbf{Supplemental material}.

\subsection{Evaluation Metrics}

We use the widely-adopted mAP metric~\cite{coco} to evaluate the detection performance of different methods. When assessing performance over corrupted data, we utilize the mean performance under corruption (mPC) metric~\cite{michaelis2019benchmarking,hendrycks2019benchmarking}. This metric represents the average mAP over various corruption types and severity levels, as expressed below:

\begin{equation}
mPC = \frac{1}{N_C} \sum_{c=1}^{N_C} \frac{1}{N_S} \sum_{s=1}^{N_S} mAP_{c,s}.
\end{equation}
where $mAP_{c,s}$ represents the performance measure evaluated on corruption type $c$ under severity level $s$, $N_c$ is the number of corruption types (15 in our work), and $N_s$ is the number of severity levels considered (5 in our work).

\begin{table}[t!]
\centering
\caption{The performance of different modality inputs on the PKU-DDD17-Car and DSEC datasets. }
\begin{tabular}{c|c|c}
\toprule[1.5pt]
Method & PKU-DDD17-Car ($mAP_{50}$ \%) & DSEC (mAP \%) \\
\midrule
Events Only & 46.5 & 12.0 \\
Frames Only & 82.7 & 25.0\\
\rowcolor{mygray}
CAFR (Ours) & \textbf{86.7} & \textbf{38.0}\\
\bottomrule[1.5pt]
\end{tabular}
\label{inputs}
\centering
\end{table}

\subsection{Ablation Study}

In this section, we conduct a series of ablation experiments to assess the effectiveness of the proposed modules. The results are summarized in Tab.~\ref{inputs} and Tab.~\ref{tab:settings}.

\fakeparagraph{Multi-modal vs. single-modal.}
In Tab.~\ref{inputs}, we compare the performance of different modality inputs. Firstly, RetinaNet~\cite{RetinaNet} is trained using events and frames as input, respectively. The results demonstrate that, despite event cameras effectively capturing dynamic semantics and filtering out redundant background information, event-based detectors exhibit inferior performance compared to their frame-based counterparts. This performance gap is attributed to the absence of crucial color and texture information for object detection. Given the complementary nature of events and frames, enhancing performance through their fusion is a viable approach. Hence, exploring effective fusion methods for events and frames is crucial. In this work, we propose CAFR to effectively leverage the complementary information from events and frames. Integrating our CAFR modules into RetinaNet, the performance can be significantly improved by $\textbf{4}\%$ and $\textbf{13}\%$ on the PKU-DDD17-Car dataset and DSEC dataset, respectively.

\begin{table}[t!]
\centering
\caption{Ablation study on key components. This table compares key components of the CAFR: ``Mul\&Add'' (multiplication and addition operations), ``CrossAtt'' (cross-self-attention), and ``Branch'' (module implementation branch). The bold value indicates the highest score.}
\begin{tabular}{cccc|c|c}
\toprule[1.5pt]
 Mul\&Add & CrossAtt & FR & Branch& $mAP_{50}$ (\%)  & mAP (\%) \\
\midrule
 \checkmark &   &   & Both & 84.3&41.5 \\
  &  \checkmark &   & Both & 84.2&42.2\\
  &   &  \checkmark & Both & 84.4&43.6 \\
 \checkmark & \checkmark &   & Both &82.7 &42.4 \\
 \checkmark &   & \checkmark & Both & 83.0&43.2 \\
 &  \checkmark  & \checkmark & Both & 86.1&45.3\\
 \midrule
 \checkmark & \checkmark & \checkmark & (a)  & 86.3& 44.7 \\
 \checkmark & \checkmark & \checkmark & (b) & 83.5& 42.1 \\
 \checkmark & \checkmark & \checkmark & Both &\textbf{86.7}& \textbf{46.0} \\
\bottomrule[1.5pt]
\end{tabular}
\label{tab:settings}
\end{table}

\begin{figure}[t!]
\centering
\includegraphics[width=0.87\linewidth]{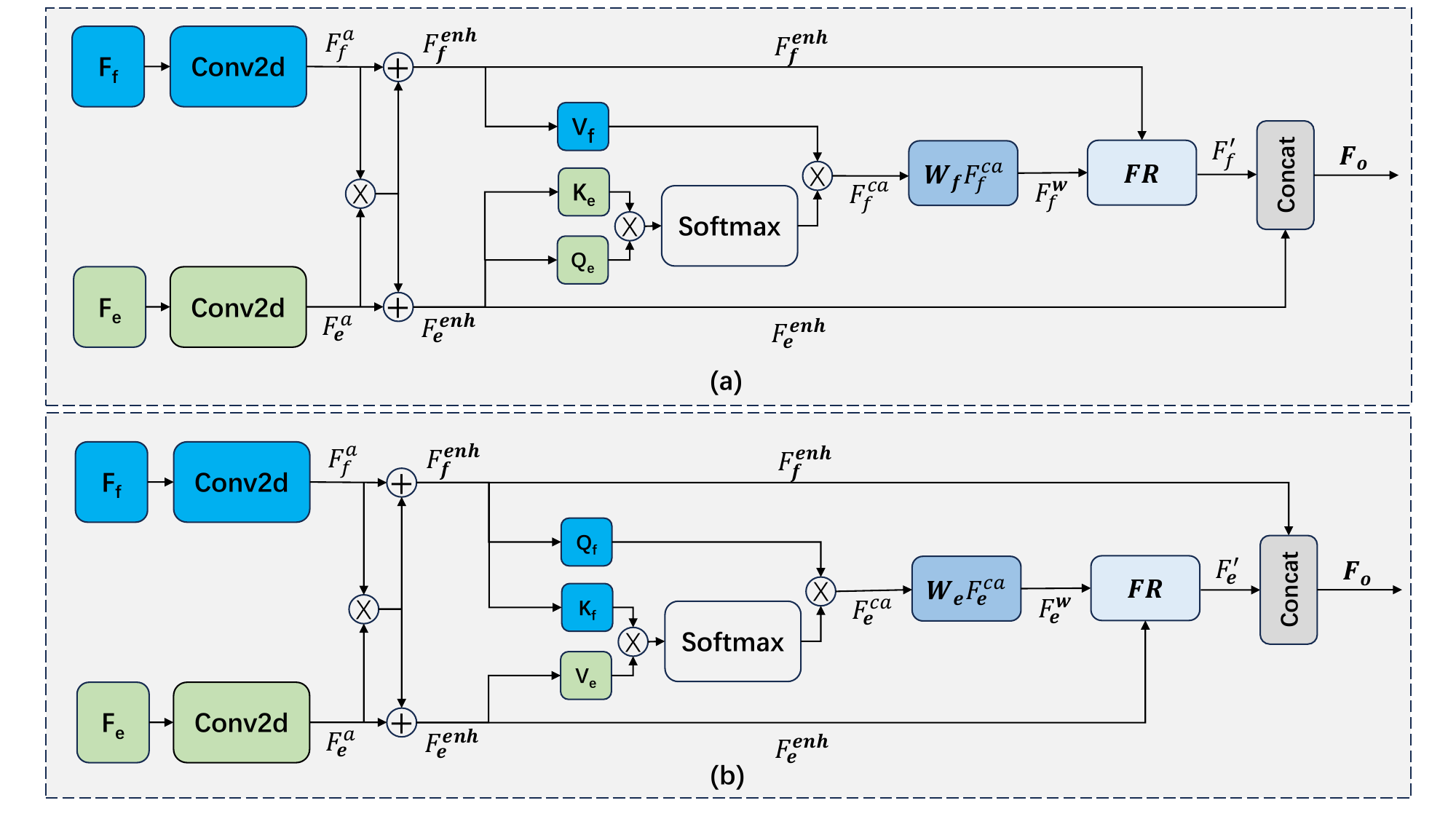}
    \caption{Different network architecture designs for the fusion module. It includes: (a) a single branch consisting of only frame-dominated CrossAtt and FR; and (b) a single branch consisting of only event-dominated CrossAtt and FR.} 
\label{CAFR_Ablation}
\end{figure}

\fakeparagraph{Effectiveness of each key component of CAFR.}
To further investigate the impact of different components on object detection, we train our model with varying network settings on the PKU-DDD17-Car dataset using events and frames as input. As shown in Tab.~\ref{tab:settings}, the experimental results demonstrate that combining all components, Mul\&Add, CrossAtt, and FR, achieves the best detection performance.

\fakeparagraph{Multi-branch vs. single-branch.}
We further explore the effect of retaining only one network branch in CAFR. The different network architecture designs are illustrated in Fig.~\ref{CAFR_Ablation}. In this figure, (a) and (b) represent a single branch consisting of only frame-dominated CrossAtt and FR and a single branch consisting of only event-dominated CrossAtt and FR, respectively. The results in Tab.~\ref{tab:settings} demonstrate that the dual network branch (as depicted in Fig.~\ref{AFRM}) achieves superior performance compared to using a single network branch. This suggests that both the event-based branch and the frame-based branch are important.

\subsection{Comparison with SOTA Methods}
We conduct a comprehensive evaluation by comparing our method with SOTA methods on both the DSEC dataset and the PKU-DDD17-Car dataset. Additionally, we thoroughly analyze the robustness of our method, particularly its effectiveness in handling corrupted data. The results showcase the strengths of our approach, demonstrating excellent performance and notable robustness in challenging scenarios. We delve into the specific details in the following:

\begin{table}[t!]
\centering
\caption{Comparison with SOTA methods on the DSEC dataset.}
\resizebox{\linewidth}{!}{ 
\begin{tabular}{c|c|c|ccc|c}
\toprule[1.5pt]
\multirow{2}{*}{Modality} & \multirow{2}{*}{Method} & \multirow{2}{*}{Model type}   & \multicolumn{4}{c}{mAP (\%)} \\
\cmidrule{4-7}
&  &&  Car &  Pedestrian& Large vehicle&Average\\
\midrule
\multirow{14}{*}{Events + Frames}
 & SENet~\cite{hu2018squeeze}&\multirow{3}{*}{Attention} & 38.4&14.9&26.0 &  26.2\\
& ECANet~\cite{wang2020eca}  & &36.7&12.8&27.5&  25.7\\
 & CBAM~\cite{woo2018cbam}&  &37.7&13.5&27.0 & 26.1\\
\cmidrule{2-7}
 & SAGate~\cite{chen2020-SAGate}  &\multirow{3}{*}{RGB-D} & 32.5&10.4&16.0&  19.6 \\
  & DCF~\cite{dcf} && 36.3&12.7&28.0&  25.7\\
  & SPNet~\cite{zhou2021specificity} & & 39.2&17.8& 26.2& 27.7\\
\cmidrule{2-7}
& FPN-Fusion~\cite{FPN_Fusion}&\multirow{8}{*}{RGB-E}& 37.5&10.9&24.9& 24.4 \\
& DRFuser~\cite{munir2023multimodal}& & 38.6&15.1&30.6&  28.1 \\
& RAMNet~\cite{gehrig2021combining}  & & 24.4&10.8& 17.6& 17.6\\
& CMX~\cite{CMX}  & & 41.6&16.4& 29.4& 29.1\\
& FAGC~\cite{FAGC}  & & 39.8&14.4& 33.6& 29.3\\
& RENet~\cite{zhou2023rgb}  & & 40.5&17.2& 30.6& 29.4\\
& EFNet~\cite{sun2022event}  & & 41.1&15.8& 32.6& 30.0\\
\rowcolor{mygray}
 & CAFR (Ours) && \textbf{49.9} & \textbf{25.8} & \textbf{38.2} & \textbf{38.0}\\
\bottomrule[1.5pt]
\end{tabular}
}
\label{DSEC}
\end{table}

\begin{table}[t!]
\centering
\caption{Comparison with SOTA methods on the PKU-DDD17-Car dataset.}
\resizebox{\linewidth}{!}{ 
\begin{tabular}{c|c|c|c|c|c}
\toprule[1.5pt]
Modality & Method  & Input representation & Model type &  $mAP_{50}$ (\%)&  mAP (\%)\\
\midrule
\multirow{2}{*}{Events} & MTC~\cite{multi_cue}& Channel image&  \multirow{2}{*}{Events only}&  47.8 &  -\\
& ASTMNet~\cite{ASTMNet}& Event embedding&&  46.2 & -\\
 \midrule
 \multirow{3}{*}{Frames}& SSD~\cite{liu2016ssd}& \multirow{3}{*}{Frame} & \multirow{3}{*}{Frames only} &  73.1 & -\\
 & Faster-RCNN~\cite{Faster-R-CNN}&  & &  80.2&  -\\
 & YOLOv4~\cite{bochkovskiy2020yolov4}&  & &  81.3&  -\\
\midrule
\multirow{15}{*}{Events + Frames} & SENet~\cite{hu2018squeeze}& \multirow{3}{*}{Voxel grid + Frame} &\multirow{3}{*}{Attention} &  81.6 & 42.4\\
& ECANet~\cite{wang2020eca}&  & &  82.2 & 40.8\\
 & CBAM~\cite{woo2018cbam}&  & &  81.9 &  42.8\\
\cmidrule{2-6}
  & SAGate~\cite{chen2020-SAGate}& \multirow{3}{*}{Voxel grid + Frame}&\multirow{3}{*}{RGB-D} &  82.0 &  43.4\\
   & DCF~\cite{dcf}&  & &  83.4&  42.5\\
     & SPNet~\cite{zhou2021specificity}&  & &  84.7&  43.3\\
     \cmidrule{2-6}
  & JDF~\cite{JDF}& Channel image + Frame &\multirow{9}{*}{RGB-E}&  84.1& -\\
  & FPN-Fusion~\cite{FPN_Fusion}& Voxel grid + Frame & & 81.9 & 41.6\\
   & DRFuser~\cite{munir2023multimodal}& Voxel grid + Frame & &  82.6 & 42.4\\
   & RAMNet~\cite{gehrig2021combining}  & Voxel grid + Frame & &  79.6 & 38.8\\
   & CMX~\cite{CMX}  &  Voxel grid + Frame & &  80.4 & 39.0\\
   & FAGC~\cite{FAGC}  & Voxel grid + Frame & &  84.8 & 42.4\\
   & RENet~\cite{zhou2023rgb}  &  Voxel grid + Frame & &  81.4 & 43.9\\
  & EFNet~\cite{sun2022event}& Voxel grid + Frame  & & 83.0 & 41.6\\
\rowcolor{mygray}
 & CAFR (Ours)&  Voxel grid + Frame && \textbf{86.7} & \textbf{46.0} \\
\bottomrule[1.5pt]
\end{tabular}
}
\label{DDD17}
\centering
\end{table}

\fakeparagraph{Comparison with SOTA methods on the DSEC dataset.}
To the best of our knowledge, there is limited research on RGB-event fusion methods specifically applied to the DSEC dataset. To address this gap, we replace our proposed fusion modules with SOTA alternatives from both the RGB-Event domain, including FPN-Fusion~\cite{FPN_Fusion}, DRFuser~\cite{munir2023multimodal}, RAMNet~\cite{gehrig2021combining}, CMX~\cite{CMX}, FAGC~\cite{FAGC}, RENet~\cite{zhou2023rgb}, and EFNet~\cite{sun2022event}, and the RGB-D domain, comprising SAGate~\cite{ADF}, DCF~\cite{dcf}, and SPNet~\cite{zhou2021specificity}. Additionally, we compare our method with well-known attention modules such as SENet~\cite{hu2018squeeze}, ECANet~\cite{wang2020eca}, and CBAM~\cite{woo2018cbam}. We maintain consistency in event representation (voxel grid), network structure (RetinaNet), loss, and hyperparameter settings across all comparative experiments, with the only variation being the fusion module. The experimental results are summarized in Tab.~\ref{DSEC}. Compared with other methods, our CAFR achieves significant improvements. Notably, CAFR outperforms the second-best method, EFNet~\cite{sun2022event}, by an impressive margin of $\textbf{8.0}\%$.

\fakeparagraph{Comparison with SOTA methods on the PKU-DDD17-Car dataset.}
Serveral works~\cite{ASTMNet,JDF,FAGC} have been conducted on the PKU-DDD17-Car dataset. Similar to DSEC, we compare our method with SOTA alternatives from the RGB-Event domain, the RGB-D domain, and attention modules. The results are listed in Tab.~\ref{DDD17}. It is evident that frame-based detectors outperform event-based detectors due to the crucial color and texture information lacking in event data. Compared with single-modal methods, fusion-based methods perform better. In particular, our CAFR consistently improves performance on both the DSEC dataset and the PKU-DDD17-Car dataset, showcasing its excellent generalization ability. Our CAFR achieves the best performance in terms of $mAP_{50}$ and $mAP$ with accuracy of 86.7\% and 46.0\%, respectively. We also visualize the detection results in challenging scenarios(See in \textbf{Supplemental Material}).

\begin{table}[t!]
\centering
\caption{The performance of different methods under various corruption conditions, including noise, blur, weather, and digital. }
\begin{tabular}{c|c|c|cccc}
\toprule[1.5pt]
\multirow{2}{*}{Method}&\multirow{2}{*}{Model type}&  \multicolumn{5}{c}{$mPC_{50}$(\%)}   \\
\cmidrule{3-7}
&&Average & Noise& Blur& Weather& Digital \\
\midrule
Frames only\cite{RetinaNet}&Frames only& 38.7& 47.6& 25.3& 28.5& 53.0 \\
\midrule
SENet~\cite{hu2018squeeze}&\multirow{3}{*}{Attention}& 63.6&68.6&56.6& 58.9& 70.3\\
ECANet~\cite{wang2020eca} & &67.1&72.6&57.6&  66.8& 71.4\\
CBAM~\cite{woo2018cbam}&& 65.2&69.9&57.2 & 62.4& 70.3\\
\midrule
SAGate~\cite{chen2020-SAGate} &\multirow{3}{*}{RGB-D} & 63.6&68.1&55.9& 61.1& 69.4 \\
DCF~\cite{dcf}& & 65.7&70.9&57.9&  62.9& 71.1\\
SPNet~\cite{zhou2021specificity}& &  66.6&70.6& 58.7& 64.8& 72.3\\
\midrule
FPN-Fusion~\cite{FPN_Fusion}&\multirow{8}{*}{RGB-E}&64.7& 70.0&56.6&63.9& 69.4\\
DRFuser~\cite{munir2023multimodal}&& 67.7&72.1&59.4& 67.8 & 71.4\\
RAMNet~\cite{gehrig2021combining}  && 53.9 &  53.5&43.3& 54.3& 64.6\\
CMX~\cite{CMX}&&64.2& 67.7&56.0&62.9& 70.2\\
FAGC~\cite{FAGC} && 52.4&62.8&38.5& 48.4 & 59.9\\
RENet~\cite{zhou2023rgb}  && 57.2 &  58.5&\textbf{72.3}& 29.9& 68.1\\
EFNet~\cite{sun2022event} && 66.4 &  67.1&58.2& 66.7& 73.4\\
\rowcolor{mygray}
CAFR (Ours)&&\textbf{69.5}& \textbf{73.6} & 57.0& \textbf{70.6}& \textbf{76.7}  \\
\bottomrule[1.5pt]
\end{tabular}
\label{robust}
\centering
\end{table}

\fakeparagraph{Robustness.}
In this experiment, we assess the model's robustness using our generated corruption data. All models undergo training with clean data and subsequent testing on the corrupted data. The corrupted data encompasses 15 diverse corruption types, as detailed earlier, broadly categorized into four groups: noise, blur, weather, and digital. Robustness evaluations are conducted across five severity levels, ranging from 1 to 5, for each corruption type. The comprehensive results are presented in Tab.~\ref{robust}. Compared to single-modal input using frames only, the fusion of events and frames leads to a substantial improvement in the model's robustness. For instance, our CAFR exhibits significantly better robustness (\textbf{69.5}\%  versus \textbf{38.7}\% ) compared to using frames only. In comparison to other fusion methods, our proposed CAFR demonstrates superior performance. These findings highlight the effectiveness of CAFR in strengthening the model against corrupted data across diverse severity levels and types.

\section{Conclusion}
This paper proposes a novel hierarchical feature refinement network for event-frame fusion. The key idea is the coarse-to-fine fusion module, named the cross-modality adaptive feature refinement module (CAFR). The CAFR employs attention mechanisms and feature statistics to enhance feature representations at the feature level. Extensive experiments are conducted on two datasets: the low-resolution PKU-DDD17-Car dataset and the high-resolution DSEC dataset. The results consistently demonstrate performance improvements on both datasets, showcasing the method's effectiveness and strong generalization. Furthermore, the model's robustness is thoroughly evaluated on generated corruption data, revealing superior robustness compared to other SOTA methods. These findings highlight the effectiveness of the proposed CAFR in strengthening the model across diverse scenarios and datasets.

\par \noindent \textbf{Acknowledgments}: This work is supported by the MANNHEIM-CeCaS (Central Car Server – Supercomputing for Automotive, No. 16ME0820), in part by National Natural Science Foundation of China (No. 62372329), in part by Shanghai Scientific Innovation Foundation (No.23DZ1203400), in part by Shanghai Rising Star Program (No.21QC1400900), in part by Tongji-Qomolo Autonomous Driving Commercial Vehicle Joint Lab Project, and in part by Xiaomi Young Talents Program.

%
%
\bibliographystyle{splncs04}
\bibliography{main}

\clearpage
\appendix
\clearpage
\setcounter{page}{1}

\section{More Experimental Details}
\label{sec:Settings}

\subsection{Datasets} 
\fakeparagraph{DDD17.}
The vehicles in the DDD17 dataset were manually labeled by the authors of~\cite{JDF} to create the PKU-DDD17-Car dataset for object detection tasks. Additional details about the PKU-DDD17-Car dataset are outlined in Tab.~\ref{tab:ddd17}.

\fakeparagraph{DSEC.}
The original dataset comprises 53 sequences captured in three distinct regions of Switzerland. RGB frames are captured using the FLIR color camera, which boasts a resolution of 1440 × 1080. The original DSEC dataset~\cite{gehrig2021dsec}  lacks the required labels for object detection. The labels introduced in ~\cite{FPN_Fusion} is used in this work. This annotated dataset consists of a total of 41 sequences, allocated for training (33 sequences), validation (3 sequences), and testing (5 sequences). Notably, the dataset covers a wide range of lighting conditions, from ideal to highly challenging. This diversity guarantees comprehensive testing of vision systems, ensuring their robustness and applicability in real-world settings. Additionally, unlike the DDD17 dataset, in which both event data and RGB data are sourced from the same DAVIS camera, the raw event data and RGB data in the DSEC dataset are not aligned; they do not correspond to the same frame. Therefore, additional preprocessing of the raw DSEC data is essential. Further details regarding the preprocessing steps are provided below:

\emph{Homographic transformation.}
To address the misalignment between the event data and RGB data, the two types of data need to be transformed into the same frame. The dataset provides a baseline of 4.5 cm between the two cameras, allowing for the transformation to a common viewpoint. In our pre-processing, we leverage the homographic transformation induced by pure rotation, as derived in Eq.~\ref{ht}, to align the scene from an RGB frame to an event-camera frame. It's important to note that, in our scenario, the scene appears far away from the camera, and the baseline of 4.5 cm is smaller than the distances of the scene objects.

\begin{table}[t!]
\centering
\caption{A detailed description of the recorded data in the PKU-DDD17-CAR dataset}	
		\begin{tabular}{c|c|c|c}
			\toprule[1.5pt]
			Recorded data (.hdf5)&Condition&Length (s) &Type \\
			\midrule
			1487339175& day & 347 &  test\\
			1487417411& day & 2096 & test \\
			1487419513& day & 1976 &  train \\
			1487424147& day& 3040&  train  \\
			1487430438& day & 3135 &  train  \\
			1487433587& night-fall & 2335 &  train  \\
			1487593224& day & 524 &  test\\
			1487594667& day & 2985 &  train  \\
			1487597945& night-fall & 50 &  test \\
			1487598202& day & 1882 &  train \\
			1487600962& day & 2143 &  test\\
			1487608147& night-fall & 1208 &  train  \\
			1487609463& night-fall & 101 &  test\\
			1487781509& night-fall & 127 &  test \\ 
			\bottomrule[1.5pt]
		\end{tabular}
\label{tab:ddd17}
\end{table}

\begin{figure}[t!]
  \centering
  \includegraphics[width=0.8\linewidth]{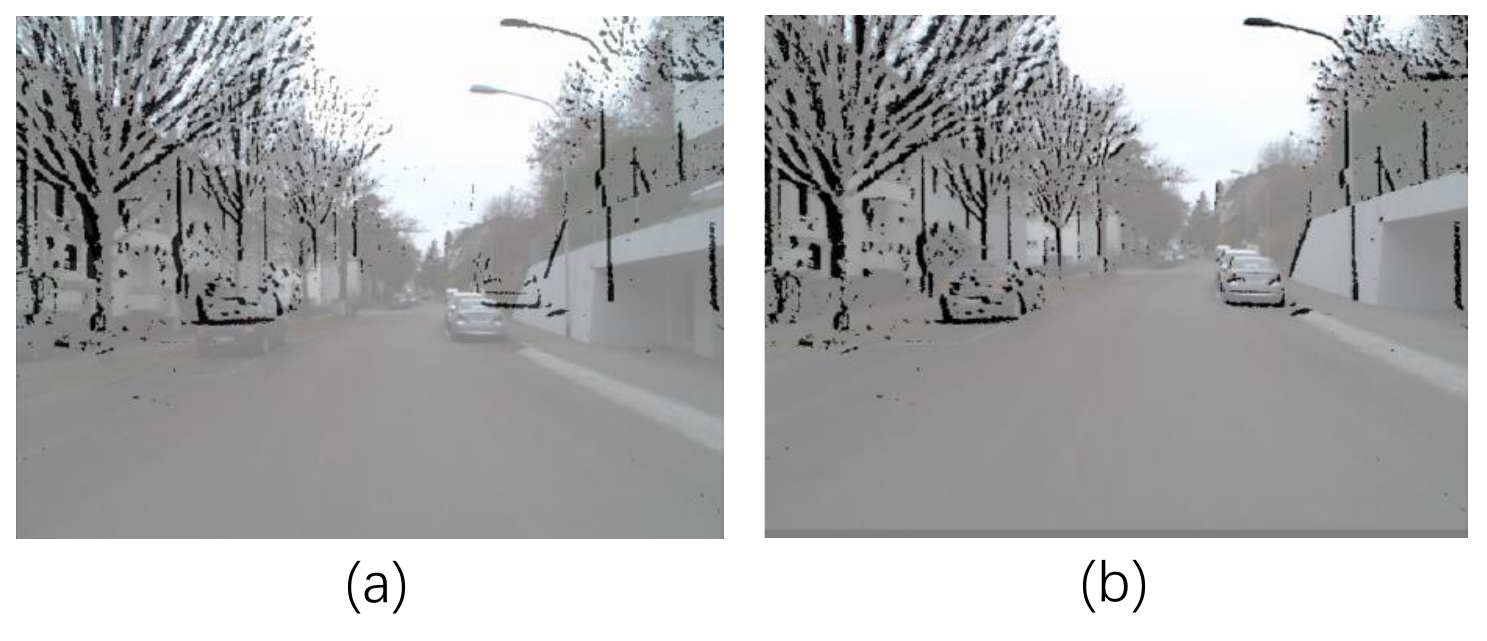}
  \hfill
  \caption{Examples illustrate the comparison before and after the homomorphic transformation. On the left (a): an overlay map of the original RGB image and event image. On the right (b): an overlay map of the RGB image and event image after the homomorphic transformation.}
  \label{fig:HT}
\end{figure}

\begin{equation}
P_{\text{event,rgb}} = K_{\text{event}} * R_{\text{rgb}} * R_{\text{event,rgb}} * R_{\text{event}}^T * K_{\text{rgb}}^{-1}.
\label{ht}
\end{equation}
where $K_{\text{rgb}}$ and $K_{\text{event}}$ represent the intrinsic camera matrices of the RGB and event cameras, respectively. Similarly, $R_{\text{rgb}}$ and $R_{\text{event}}$ denote the rotation matrices accounting for the transition from distorted to undistorted frames for each camera. Additionally, $R_{\text{event,rgb}}$ stands for the rotation matrix aligning the RGB camera coordinate system with that of the event camera.

Fig.~\ref{fig:HT} illustrates the impact of the homomorphic transformation. In the left figure, the RGB image and the event image are not aligned. However, after applying the homomorphic transformation, the RGB image and the event image become aligned, sharing the same frame and being suitable for multimodal fusion. Additionally, the size of the RGB images is resized to match the event image, ensuring that both types of data have the same field of view and resolution.

\begin{table}[t!]
    \centering
    \caption{Object annotations in the labeled DSEC dataset.}
    \begin{tabular}{cccc}
    \toprule[1.5pt]
    Categories & Car & Pedestrian & Large vehicle (Bus \& Truck) \\
    \midrule
    Count & 100068 & 17126 & 14771 \\
    Percentage & 0.76 & 0.13 & 0.11 \\
    \bottomrule[1.5pt]
    \end{tabular}
    \label{obejctanno}
\end{table}

\begin{table}[t!]
    \centering
    \caption{Data amount in the labeled DSEC dataset.}
    \begin{tabular}{ccccc}
    \toprule[1.5pt]
    Type & Train & Val & Test & Total \\
    \midrule
    Sequences & 33 & 3 & 5 & 41 \\
    Frames & 44148 & 3642 & 4896 & 52686 \\
    \bottomrule[1.5pt]
    \end{tabular}
    \label{tab:DSEC-data}
\end{table}

\begin{figure}[t!]
  \centering
  \includegraphics[width=0.8\linewidth]{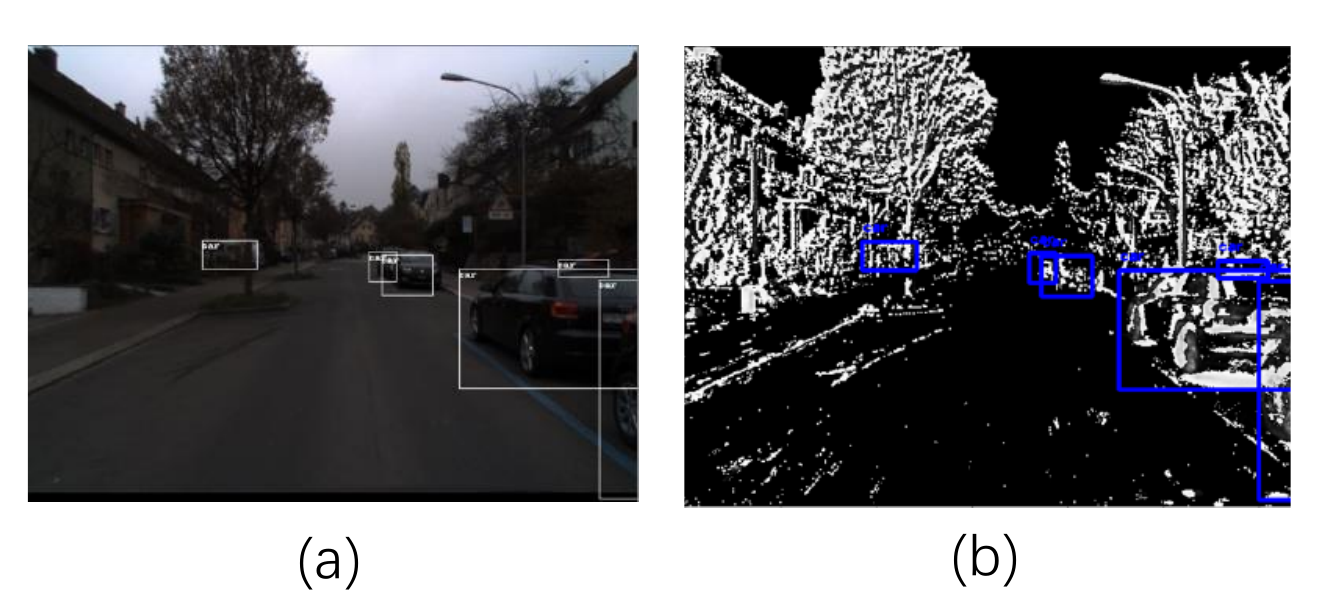}
  \hfill
  \caption{Examples of events and RGB images with annotations. Left (a): RGB image; Right (b): event frame.}
  \label{fig:label}
\end{figure}

\emph{Annotation generation.}
To facilitate object detection on the DSEC dataset, we employed simulated annotations provided by \cite{FPN_Fusion}. YOLOv5 \cite{yolo5} was used to label RGB images. The homographic transformation was applied to transfer labels from RGB images to the event frame, accounting for objects within the event camera's 640 × 480 resolution. Examples of labeled events and RGB images are depicted in Fig.~\ref{fig:label}, encompassing three object classes: car, pedestrian, and large vehicle (refer to Tab.~\ref{obejctanno}).

After the aforementioned pre-processing steps, we acquired the labeled DSEC dataset for experimentation. The dataset comprises a total of 41 sequences, with 33 sequences allocated for training, 3 sequences for validation, and the remaining 5 sequences designated for testing. Detailed information about the data is presented in Tab.~\ref{tab:DSEC-data}, where the event data file is approximately 300 GB and the RGB data file is around 23 GB.

\begin{table}[t!]
\centering
\caption{The details of different corruption types. }
\begin{tabular}{c|c}
\toprule[1.5pt]
Group & Corruption Type \\
\midrule
Noise & Gaussian Noise, Shot Noise, Impulse Noise \\
Blur& Defocus Blur, Glass Blur, Motion Blur, Zoom Blur\\
Weather & Fog, Snow, Frost, Brightness \\
Digital & Contrast, Elastic Transform, Pixelate, Jpeg Compression \\
\bottomrule[1.5pt]
\end{tabular}
\label{corruption_types}
\end{table}

\begin{figure*}[t!]
  \centering
  \includegraphics[width=0.9\linewidth]{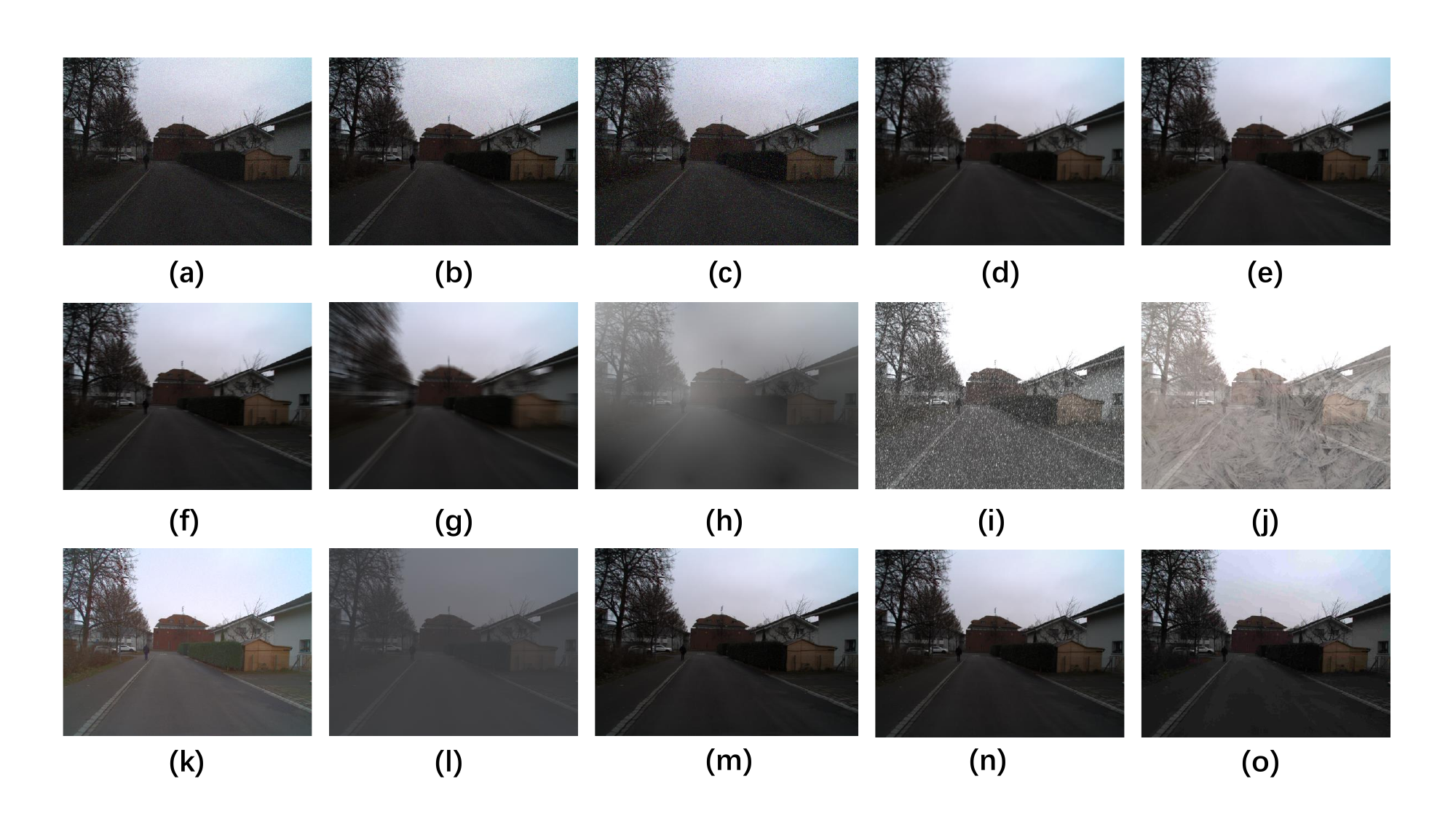}
  \hfill
  \caption{The dataset encompasses 15 types of algorithmically generated corruptions, categorized into noise, blur, weather, and digital groups. The corrupted images, denoted from (a) to (o), include Gaussian Noise, Shot Noise, Impulse Noise, Defocus Blur, Glass Blur, Motion Blur, Zoom Blur, Fog, Snow, Frost, Brightness, Contrast, Elastic Transform, Pixelate, and Jpeg Compression. Each corruption type exhibits five severity levels, resulting in a total of 75 distinct corruptions. The images presented here correspond to severity level 2.}
  \label{fig:corruption}
\end{figure*}

\fakeparagraph{Corruption data.}
In this work, we introduced 15 types of corruption, each with five levels of severity, to assess the impact of diverse corruption types on object detection models. The chosen corruption types are categorized into four groups: noise, blur, weather, and digital. The specific corruption types are detailed in Tab.~\ref{corruption_types}. Fig.~\ref{fig:corruption} provides an illustration of these corruption types at severity level 2. All corruption treatments are applied to the test set, enabling us to evaluate a model's robustness against previously unseen corruptions.

\emph{Noise.} The first corruption type is Gaussian noise. This corruption may occur in low-light conditions. Electronic noise caused by the discontinuous character of light is known as shot noise, also referred to as poisson noise. Impulse noise is the color analog of salt-and-pepper noise and can be caused by bit errors.

\emph{Blur.} Defocus blur occurs when the image is out of focus. Glass blur appears on "frosted glass" windows or panels. When the camera is moving swiftly, motion blur happens. When the camera advances quickly toward an item, zoom blur occurs.

\emph{Weather.} Snow is a type of precipitation that impairs vision. When there are ice crystals on the lenses or windows, frost occurs. A diamond-square method is used to render the fog that surrounds the items. The brightness varies with the intensity of daylight.

\emph{Digital.} Based on the lighting and the subject’s color, contrast can be either high or low. The elastic transform enlarges or reduces picture regions. Pixelate occurs when upsampling low-resolution images. Jpeg compression is a lossy image compression format that produces compression artifacts.

\begin{figure}[t!]
  \centering
  \includegraphics[width=0.5\linewidth]{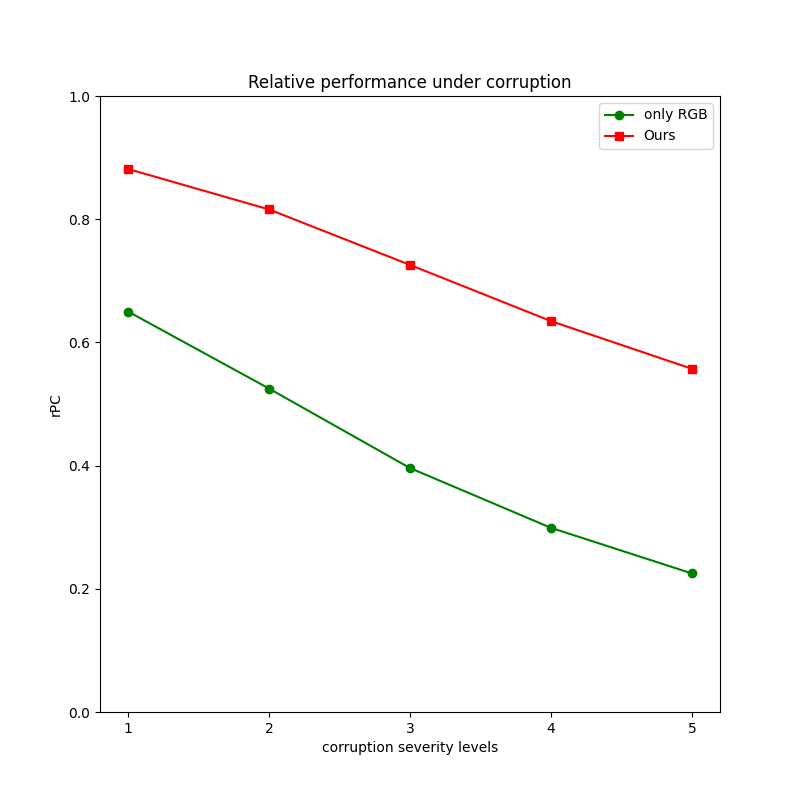}
  \hfill
  \caption{Relative performance under various severity.}
  \label{fig:cures}
\end{figure}

\subsection{Training details.} 
Our model is implemented using PyTorch~\cite{PyTorch}, and we initialize the event-based and frame-based backbone branches with pre-trained ResNet-50~\cite{resnet}.
The input data for the PKU-DDD17-Car dataset and DSEC dataset are resized to $346\times 260$ and $640\times 480$, respectively. Furthermore, we train the model using the Adam optimizer~\cite{kingma2014adam} with an initial learning rate of  \(1 \times 10^{-4}\).
The experiments are carried out on an Nvidia RTX 3090 GPU with a system running Ubuntu 20.04. During training, the input from the RGB image is set to blank (zero) with a 15\% probability. This strategy compels the sensor fusion model to extract information primarily from the second modality, the event camera, thereby enhancing the model's robustness against corruption in the frame-based camera. The total number of epochs is set to 200, with a batch size of 8 for the PKU-DDD17-Car dataset and 1 for the DSEC dataset, respectively.

\section{More Experimental Analysis}

\fakeparagraph{Performance under different lighting conditions.}
To analyze the contribution of the event camera, we assess the performance gain under different illumination conditions. Detailed comparisons are provided in Tab.~\ref{light}. The results illustrate that our proposed CAFR, capitalizing on the complementary nature of events and frames, consistently improves detection performance and is better than other fusion methods across various lighting conditions. This analysis provides valuable insights into the effectiveness of incorporating event data in improving overall model performance under diverse illumination scenarios.

\begin{table}[t!]
        \centering
        \caption{Comparison with SOTA fusion alternatives: COCO mAP@0.50:0.95 on the PKUDDD17-CAR dataset for the different methods.}
            \begin{tabular}{c| c|c| c c| c}
                \toprule
                Method & Model type&Test(All day)& Test(Day) & Test(Night)& FPS \\
                \midrule
                SENet~\cite{hu2018squeeze} &\multirow{3}{*}{Attention}& 42.4 & 43.7 & 37.0 & 8.0 \\
                ECA~\cite{wang2020eca} && 40.8 & 42.2 & 36.1 & 7.6 \\
                CBAM~\cite{woo2018cbam} & &42.8 & 44.2 & 38.0 & 10.3 \\
                \midrule
                SAGate~\cite{chen2020-SAGate} &\multirow{3}{*}{RGB-D}& 43.4 & 44.9 & 38.0 & 11.8 \\
                DCF~\cite{dcf}& & 42.5 & 43.4 & 39.0 & 13.8 \\
                SPNet~\cite{zhou2021specificity} && 43.3 & 44.9 & 37.1 & 9.1 \\
                \midrule
                FPN-Fusion~\cite{FPN_Fusion}&\multirow{8}{*}{RGB-E}&41.6& 43.2&35.7&12.0\\
                DRFuser~\cite{munir2023multimodal} && 42.4 & 43.3 & 38.8 & 11.5 \\
                RAMNet~\cite{gehrig2021combining} && 38.8 & 39.2 & 36.9 & 11.5 \\
                CMX~\cite{CMX}& & 39.0 & 40.2 & 35.4 & 2.8 \\
                FAGC~\cite{FAGC} & &42.4 & 43.7 & 36.7 & 5.3 \\
                RENet~\cite{zhou2023rgb} && 43.9 & 45.4 & 39.1 & 5.0 \\
                EFNet~\cite{sun2022event} && 41.6 & 43.4 & 35.1 & 9.7 \\
                CAFR (Ours)& &\textbf{ 46.0} & \textbf{46.9} & \textbf{42.1} & 6.4\\
                \bottomrule
            \end{tabular}
        \label{light}
\end{table}

\fakeparagraph{Efficiency analysis.}
As detailed in Tab.~\ref{light}, the running speeds of various methods are presented. 
In comparison with other fusion methods, such as CBAM~\cite{woo2018cbam}, SAGate~\cite{chen2020-SAGate}, and RENet~\cite{zhou2023rgb}, the proposed method exhibits comparable running speed while significantly enhancing detection accuracy.

\fakeparagraph{More robustness analysis.} 
Fig.~\ref{fig:cures} illustrates the relative performance under corruption (RPC) at severity levels ranging from 1 to 5. Across all models, there is a consistent decline in relative performance as the severity of corruption increases. Notably, the model relying solely on RGB data exhibits the steepest decline, indicating its lower robustness. In contrast, the proposed fusion method significantly improves robustness across different severity levels.

\begin{figure}[t!]
  \centering
  \includegraphics[width=0.9\linewidth]{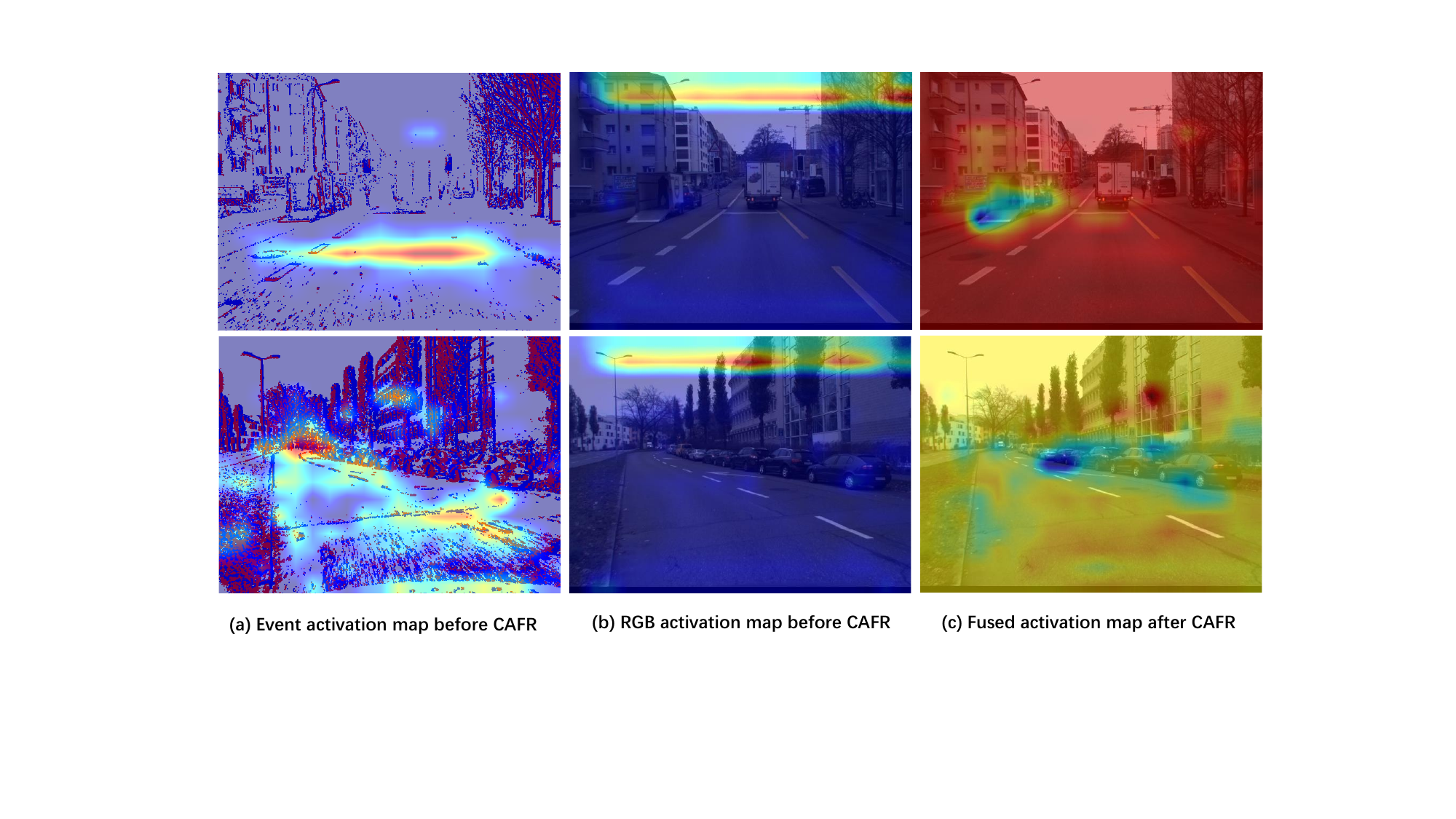}
  \caption{Representative examples of different activation maps on the DSEC dataset are: (a) event activation map before CAFR; (b) RGB activation map before CAFR; and (c) fused activation map after CAFR.}
  \label{fig:map}
\end{figure}

\fakeparagraph{Visualization of activation maps.} In the fig.~\ref{fig:map}, we visualize the activation maps of RGB and event modalities before and after CAFR. After applying CAFR, the model demonstrates enhanced focus on significant regions.

\begin{figure*}[t!]
  \centering
  \includegraphics[width=0.9\linewidth]{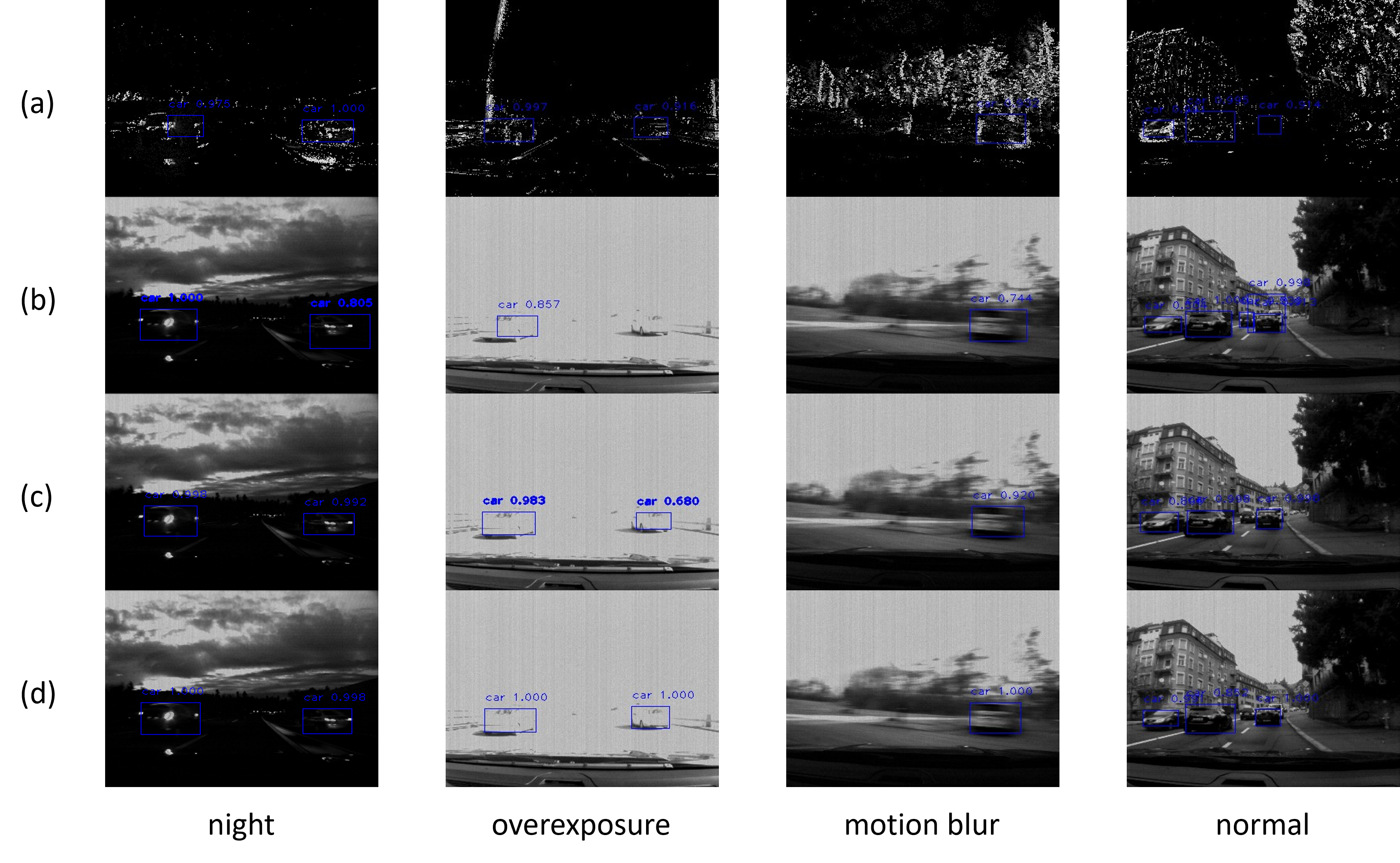}
  \hfill
  \caption{Representative examples of different object detection results on the PKU-DDD17-Car dataset: (a) our baseline using event images; (b) our baseline using frames; (c) SPNet~\cite{zhou2021specificity} (the second-best model in terms of $mAP_{50}$\% and mAP\%) using frames and events. (d) our method using frames and events.}
  \label{fig:detection}
\end{figure*}

\begin{figure*}[t!]
  \centering
  \includegraphics[width=0.9\linewidth]{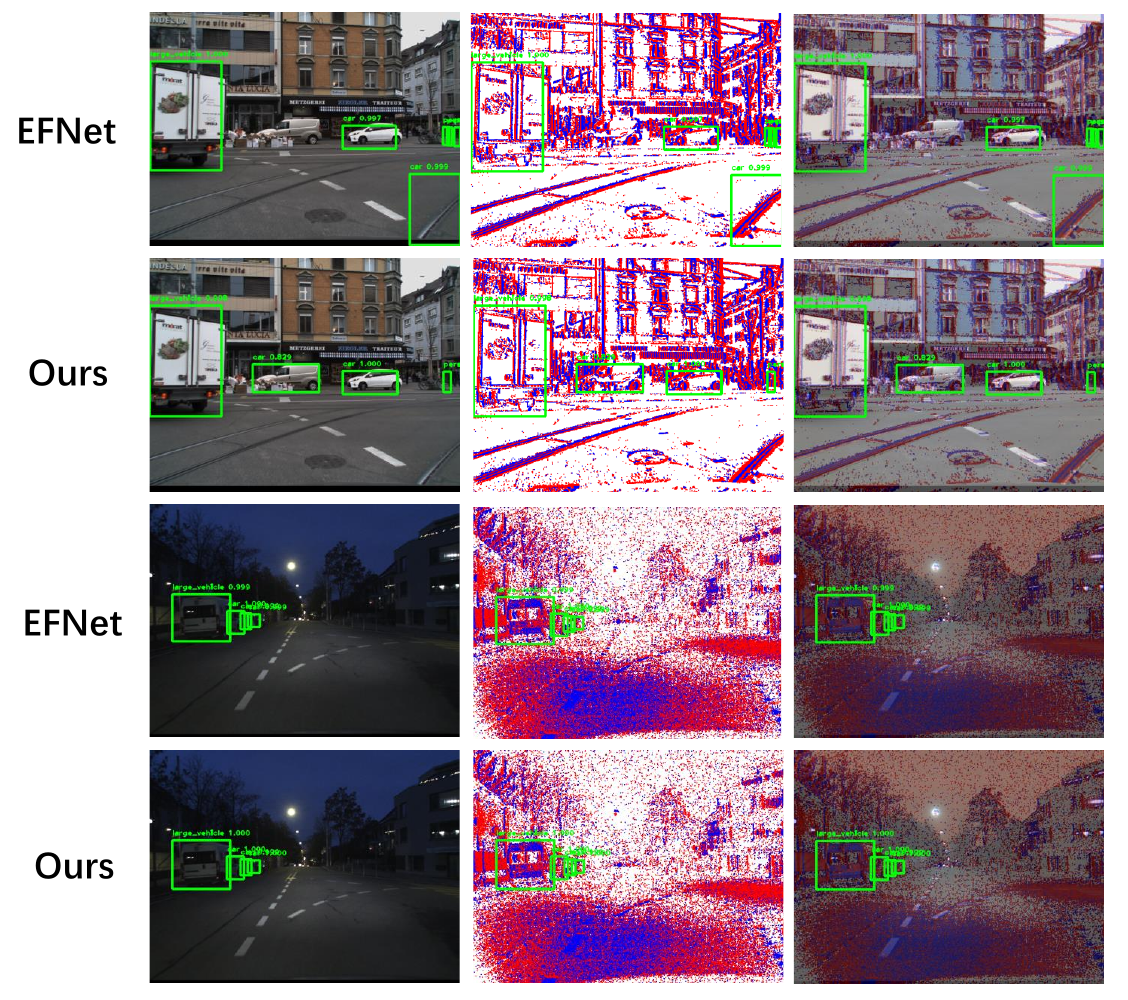}
  \hfill
  \caption{Representative examples of different object detection results on the DSEC dataset. The first two rows are daytime scenes, and the last two rows are nighttime scenes. Each column represents RGB images, event frames, and merged RGB event frames, respectively.}
  \label{fig:dsec}
\end{figure*}

\fakeparagraph{Results on detection predictions.}
In Fig.~\ref{fig:detection} and Fig.~\ref{fig:dsec}, we visualize the detection results selected from the PKU-DDD17-Car dataset and DSEC dataset, respectively. The results demonstrate that the proposed method can consistently produce satisfactory detection results in various challenging scenarios. Compared with the second-best methods, SPNet~\cite{zhou2021specificity} and EFNet~\cite{sun2022event}, our method performs better prediction results.

\section{Limitations}
While our CAFR has demonstrated efficacy in the context of object detection, it is essential to acknowledge a current limitation. The scope of our evaluations has been confined to this specific task, and extrapolating the performance of CAFR to other prevalent perception tasks, such as semantic segmentation, depth prediction, and steering angle prediction, remains unexplored. Future investigations could delve into the broader applicability of CAFR, providing insights into its adaptability and potential limitations across diverse perception domains.

\section{More Related Works}
In other areas of cross-modal fusion, such as depth, thermal, and event data, we discuss relevant works below. In the field of RGB-D salient object detection (SOD), the joint learning and densely cooperative fusion framework introduced in \cite{fu2020jl} aims to improve performance. The authors of \cite{luo2020cascade} developed graph-based techniques to design network architectures for RGB-D SOD. Additionally, an automatic architecture search approach for RGB-D SOD is presented in \cite{sun2021deep}. For semantic segmentation, the authors of \cite{seichter2021efficient} proposed the efficient scene analysis network (ESANet) for RGB-D semantic segmentation, utilizing channel attention \cite{hu2018squeeze} for RGB-D fusion. Furthermore, an uncertainty-aware self-attention mechanism is employed in \cite{ying2022uctnet} for indoor RGB-D semantic segmentation. The adaptive-weighted bi-directional modality difference reduction network proposed in \cite{zhang2021abmdrnet} addresses RGB-T semantic segmentation. Recently, a multi-modal fusion network (EISNet) introduced in \cite{10477577} aims to enhance semantic segmentation performance using events and images.

\end{document}